\documentclass{Interspeech}
\usepackage{balance}
\usepackage{placeins} 
\usepackage{url}
\usepackage{dblfloatfix}
\usepackage{multirow} 
\usepackage[dvipsnames]{xcolor}

\interspeechcameraready

\DeclareUrlCommand\surl{\urlstyle{tt}}

\title{Accurate, fast, cheap: Choose three. Replacing Multi-Head-Attention with Bidirectional Recurrent Attention for Long-Form ASR}

\author{Martin}{Ratajczak}
\author{Jean-Philippe}{Robichaud}
\author{Jennifer}{Drexler Fox}

\affiliation{}{Rev.com}{USA}
\email{martin.ratajczak@gmail.com, jp@rev.com, jennifer.drexler@rev.com}

\usepackage{comment}

\begin{document}

\maketitle
\begin{abstract}

Long-form speech recognition is an application area of increasing research focus. ASR models based on multi-head attention (MHA) are ill-suited to long-form ASR because of their quadratic complexity in sequence length. We build on recent work that has investigated linear complexity \textit{recurrent attention} (RA) layers for ASR. We find that bidirectional RA layers can match the accuracy of MHA for both short- and long-form applications. We present a strong limited-context attention (LCA) baseline, and show that RA layers are just as accurate while being more efficient. We develop a long-form training paradigm which further improves RA performance, leading to better accuracy than LCA with 44\% higher throughput. We also present \textit{Direction Dropout}, a novel regularization method that improves accuracy, provides fine-grained control of the accuracy/throughput trade-off of bidirectional RA, and enables a new \textit{alternating directions} decoding mode with even higher throughput.
\end{abstract}

\section{Introduction}


Improved ASR accuracy enables long-form audio applications (e.g., multiple hours). However, Transformers, especially their multi-head attention (MHA) component, are ill-suited for long-form ASR due to quadratic time/memory complexity in sequence length. Decoding long audio in chunks (e.g., Whisper \cite{radford2022robustspeechrecognitionlargescale} uses 30s segments) loses crucial full-audio context.

Recently, several layer types have been introduced that mimic the properties of MHA while having linear time and memory complexity in sequence length. These layers achieve linear complexity through the use of recurrent-style computations that depend only on a hidden state containing the accumulated history at each time-step; we will thus collectively refer to them here as recurrent attention (RA) layers. In this paper, we investigate the use of the RWKV \cite{peng2023rwkv,peng2024eagle} and Mamba \cite{gu2023mamba,dao2024transformers} layer types in place of MHA for long-form ASR.

When processing sequences, MHA sees the entire context at once and can take advantage of the information available both before and after the current frame. Recurrent models, by contrast, only see previous context and cannot look into the future. In this paper, we use bidirectional RA to account for this discrepancy and provide future context to the RA layers. 

We also introduce Direction Dropout (DirDrop), a novel regularization method that also enables multiple alternative inference modes, giving full control over the speed/accuracy trade-off of these models. 




\newpage
\noindent The main contributions of this paper are:
\begin{enumerate}
        \item Bidirectional RA layers (Mamba-2, RWKV v6) are competitive with MHA in Conformer encoders for ASR. Bidirectional RWKV-Conformer is more efficient than standard Conformer and limited-context attention with global tokens \cite{koluguri2024investigating} for long-form ASR at similar accuracy.
    \item With additional long-form fine-tuning, the bidirectional RWKV-Conformer is more accurate than limited-context MHA with 44\% higher throughput, while the unidirectional RWKV-Conformer is as accurate as limited-context MHA with 72\% higher throughput. 
 
    \item DirDrop regularization closes the bidirectional RA vs. MHA gap. DirDrop-trained bidirectional models achieve unidirectional inference accuracy similar to unidirectional models, unlike standard bidirectional models which fail catastrophically in unidirectional inference. In this way, our approach enables one model to be used flexibly for both offline and streaming applications.
  \item Finally, \textit{Alternating direction} decoding with DirDrop nears bidirectional decoding accuracy at unidirectional cost. This offers new insights into bidirectional information propagation: it not only shows the importance of merging information from the past and future, but also reveals that it is not necessary to process both directions within each layer.
\end{enumerate}

\section{Related Work}

Different types of recurrent attention (RA) layers have been introduced. These broadly fall into two classes: RWKV \cite{peng2023rwkv} and state-space models (SSM), which include S4 \cite{gu2021efficiently}, H3 \cite{fu2022hungry}, and Mamba \cite{gu2023mamba}. 
For this paper, we compare RWKV v6 \cite{peng2024eagle} 
and Mamba-2 \cite{dao2024transformers}. 
For more details, see Section \ref{arch_RA}.

RA layers have previously been applied to ASR, with promising results: RA layers can effectively replace MHA in the encoder \cite{fathullah2023multi, masuyama2024mamba} or self-attention in the decoder \cite{miyazaki2023structured}, for streaming recognition \cite{an2023exploring,shan2024augmenting}, and for long-form inference \cite{honda2024efficient,miyazaki2024exploring}. Our work differs from prior work on long-form ASR with RA layers in a number of ways. First, both \cite{honda2024efficient} and \cite{miyazaki2024exploring} use variants of SSM models; we compare an SSM, Mamba-2, with RWKV. Second, these prior works have used either CTC or attention-encoder decoder architectures; we use RNN-T, which has been previously shown to be best suited to long-form ASR \cite{fox2024updated}. We also use a stronger baseline and analyze model throughput in addition to accuracy.

Our Direction Dropout technique is inspired by methods that perturb the model structure~\cite{maaten2013,wager2013,bachman2014,ratajczak_2015}. These methods can be interpreted as a mixture of models and have regularization effects such as the classical Dropout and Dropconnect~\cite{hinton2012,wan2013}. It bears some similarity to LayerDrop \cite{fanreducing}. In that work, whole layers are randomly dropped during training, which both has a regularizing effect on the full model and allows the model to perform well with some layers removed at inference time. LayerDrop has been applied to ASR, with the goal of trading-off accuracy and latency depending on device-specific resource constraints \cite{shi2021dynamic}. Our work differs from this previous work in that we only drop directions of RA layers, rather than complete layers.

\section{Methods}
Our models use the Conformer-Transducer architecture that consists of a Conformer encoder \cite{gulati2020} and a Transducer decoder \cite{graves2012}. 
For all models in this paper, we maintain the same overall architecture, the same Conformer layer structure, and the same parameters for all convolutional and linear layers. The only difference is the attention computation in the encoder. 


\subsection{Replacing MHA with Recurrent Attention}\label{arch_RA}

A key component of the standard Conformer architecture for ASR is the multi-head attention (MHA) layer. MHA involves a large matrix multiplication that calculates the attention that each frame in the input sequence pays to each other frame, leading to quadratic complexity in sequence length. \textit{Recurrent attention} (RA) layers, by contrast, compute the output at each frame based on a hidden state that aggregates information from all previous frames, leading to linear complexity in sequence length. RA layers differ from the previous generation of recurrent neural networks (RNNs) through the use of new algorithms for computing all of the outputs for one layer in parallel. These algorithms enable efficient training of deep models on GPUs as well as efficient inference when the entire input sequence is available, as is the case for the ASR encoder during offline inference. 

In this paper, we investigate replacing MHA in the Conformer encoder with two variants of RA. Mamba-2 is a popular variant of a class of models called state space sequence models (SSM) that are inspired by classical state space models, but take the form of a neural network layer. Mamba-2 is augmented with a selection mechanism and carefully designed to combine the SSM attention calculation with convolution and normalization layers. For this reason, we chose to replace MHA with the entire Mamba-2 block. RWKV, by contrast, is implemented in two discrete stages: a time-mixing component that performs recurrent attention and a channel-mixing component designed to capture local transformations. We used only the time-mixing component of RWKV as a replacement for MHA, hypothesizing that the feed-forward and convolutional operations in the Conformer would effectively capture local information. 

When choosing parameters, we tried to keep the comparison between MHA, Mamba-2, and RWKV as close as possible. We use the same attention dimensions for all models. While some research suggests that Mamba models should have twice the number of layers as MHA models \cite{gu2023mamba}, we always compare models with the same number of layers. Mamba-2 can be formulated as a multi-head operation, analogous to multi-head attention; we use the same number of attention heads for Mamba-2 and MHA. 
\subsection{Limited Context Attention with Global Tokens}
An alternative mechanism to reduce the memory and computational complexity of MHA is to limit the size of the context window seen by the attention mechanism. This is the approach taken in FastConformer \cite{rekesh2023fast}, which has been shown to be both accurate and efficient for long-form ASR \cite{koluguri2024investigating}. Limited context attention (LCA) is most accurate when combined with global tokens (GT) that attend to the entire sequence. The fundamental computations that underlie LCA+GT are the same as MHA, but because the number of global tokens and the size of the context window are fixed, the attention computation is linear in the sequence length. We follow \cite{koluguri2024investigating} and train base models with full-context attention, followed by a small amount of LCA+GT fine-tuning. We use LCA+GT as a strong baseline for our long-form ASR experiments. 

\subsection{Long-Form Training}

Prior work has shown that transducer models generalize reasonably well from training on short utterances to inference on longer utterances but that training on long utterances can further improve long-form ASR performance \cite{fox2024updated}. We took the simple approach of creating longer utterances by concatenating neighboring training utterances together. When doing standard training on these concatenated utterances, we found that our transducer decoder implementation is limited to still quite short training utterances because of memory requirements. For this reason, we developed a second long-form training variant, in which we fine-tune only the attention blocks in the encoder component of the long-form trained model.

\subsection{Bidirectional RA}
RA layers have one main drawback, relative to MHA, in the context of an ASR encoder. MHA has access to all frames in the sequence at all times, while the recurrent attention at each frame only has access to information from previous frames. 
To address this mismatch, we implemented bidirectional versions of our RA layers inspired by classical bidirectional RNNs~\cite{schuster1997bidirectional}. At each layer, we add a second set of attention weights, which are used to process a reversed version of the input sequence. The outputs of this backwards pass are reversed again and then averaged with the outputs of the forward pass.

\subsection{Direction Dropout}
RA is appealing for long-form ASR due to its efficiency. While bidirectional RA is faster than LCA, it remains slower than unidirectional RA. To balance speed and accuracy, we introduce Direction Dropout (DirDrop). For each Conformer block, DirDrop randomly drops one of the RA directions with a set probability. We use two variants: DirDrop-R2L only drops right-to-left (R2L) while DirDrop-Both randomly drops either direction. Models trained with DirDrop enable flexible inference with any number of bidirectional layers, optimizing the speed/accuracy trade-off. They also support alternating directions decoding, aggregating past and future information while using only one RA direction per layer.


\section{Experiment Details}

\subsection{Model Architecture}

Our experiments are performed using the WeNet framework (see \cite{yao2021wenet}, \cite{zhang2022wenet}).  All ASR models used in this paper have the same basic architecture: a Conformer encoder, a CTC projection layer at the encoder output, a bidirectional self-attention decoder (SAD), and an LSTM-based transducer (RNN-T). This architecture gives us a large amount of flexibility when it comes to inference - each trained model has a number of decoding modes, each with their own strengths and weaknesses. For simplicity, unless otherwise noted, we use joint CTC and RNN-T decoding for all experiments in this paper, with weights of 0.3 and 0.7 for CTC and RNN-T, respectively.

All models used in this paper have 12 encoder layers. The attention size (regardless of attention type) is 512 and the size of the feed-forward layers is 2048. MHA and Mamba-2 models both have 8 attention heads. The receptance field of the RWKV models is set to 2048. For LCA+GT, we use a context window of 256 and one global token, matching the settings in \cite{koluguri2024investigating}. We used the LCA+GT implementation from NeMo\footnote{~\surl{github.com/NVIDIA/NeMo/} - the RelPositionMultiHeadAttentionLongformer class}, integrated into WeNet. For DirDrop, we use a dropout rate of 20\%. 

We used the Adam optimizer, with dynamic batching of 20k frames, a learning rate of 5e-4, 50k warmup steps and a standard weighted sum of the CTC, SAD and transducer losses.  The configuration files as well as the source code will be available online after publication\footnote{~\surl{github.com/revdotcom/paper_accurate_fast_cheap/}}.

We reused the GigaSpeech trained SentencePiece BPE tokenizer available in WeNet, which is a unigram model with a vocabulary size of 5k tokens\footnote{~\surl{github.com/wenet-e2e/wenet/tree/main/examples/gigaspeech/s0}}.
\subsection{Data}

All models are trained with the GigaSpeech XL data set, containing 10k hours of audio \cite{DBLP:journals/corr/abs-2004-05150}.
For our initial experiments, labeled as ``SF" (short-form), we performed training on the standard segments released with the dataset, which have an average length of 4.4 seconds. For long-form training (``LF" models), the training utterances used were between 10 and 15 seconds, with an average of 11.4 seconds. For long-form fine-tuning of the attention weights (``LFXL" models), the average segment length was 78.6 seconds, with a range of 75 and 85 seconds.


We use word error rate (WER) as our measure of ASR accuracy. All WER results in this paper are reported as ``dev WER/test WER". For all WER calculations, we use the filtering rules provided with the GigaSpeech dataset. These rules remove punctuation marks, filler words and hyphens from both the reference text and the ASR output.

The GigaSpeech dev and test partitions are fully transcribed - although the text is provided in small segments, the entire content of each audio file can be reconstructed by simply concatenating these segments together \cite{fox2024updated}.
We perform long-form inference on these files with a very simple chunked decoding process: we split each original full-length audio file into fixed segments of length \textit{chunk-size} and decode each one, without overlap. Chunk size is measured in frames, which are extracted from the audio at a rate of 100 frames per second.
We join these chunk outputs together in order and evaluate the full-file WER with FSTalign \cite{Del_Rio_2021}, again using the filtering rules.


\subsection{RTF/Throughput Evaluation}

We measure performance of our ASR models using an NVIDIA A10G with 24GB of VRAM. 
Throughput measurements reflect performance on a single GPU. 
One audio file of 9857 seconds was used for all reported results. 
For readability, we report the throughput as minutes of audio processed per second (MPS) instead of the usual \textit{real-time factor}. 

Since our models differ only in the attention computations in the encoder, we removed the transducer portion of the model for this part of the evaluation. 
This allows us to highlight the gain that replacing MHA can bring to the part of the model it impacts.  
We evaluated different batch sizes, from 1 to 12, but we report only on size 4 as it allowed the best throughput without sacrificing accuracy for all models tested given the selected hardware.
In each case, the model was sent two warm-up queries to remove the impact of startup time in the numbers. 
Multiple runs of the decoding task for each configuration (model, batch size, chunk size) were performed to estimate performance stability, but the variance was not significant (less than 0.01 min/s), so we report without it for clarity.


\section{Results and Discussion}

\subsection{Short-Form ASR}
\begin{table}[th]
  \caption{MHA vs. RA. WER (\%) on short-form segments.}
  \label{tab:sf_results}
  \centering
  \begin{tabular}{cccc}
    \toprule
    \textbf{Bidir.} & \textbf{MHA} & \textbf{Mamba-2} & \textbf {RWKV} \\
    \midrule
    N & -         & 11.2/11.4 & 11.2/11.3 \\
    Y & 10.9/10.9 & 11.0/11.0 & 11.0/11.0 \\
    \bottomrule
  \end{tabular}
\end{table}

Table \ref{tab:sf_results} shows results for short-form decoding. When used unidirectionally, Mamba and RWKV underperform MHA, but match it when used bidirectionally. Given similar accuracy, we focus on RWKV for the rest of the paper.


\subsection{Long-Form ASR}
\begin{table}[h]
\caption{WER (\%) performance of SF- and LF-trained models on LF decoding, as a function of decoding chunk size in frames.}
\label{tab:sf_vs_lf_training}
\centering
\begin{tabular}{c|c|c|c|c}
\toprule
    \multirow{2}{*}{\textbf{chunk}}  &
    \multicolumn{2}{c|}{\textbf{MHA}}  &
    \multicolumn{2}{c}{\textbf{bi-RWKV}}  \\ 
      & SF & LF & SF & LF \\ 
     \midrule
     2k & 13.5/14.5 & 13.3/14.2 & 13.4/14.4 & 13.2/14.1 \\ 
     9k & 13.2/14.1 & 12.5/13.4 & 12.9/13.7 & 12.6/13.4 \\ 
    20k & 15.3/16.8 & 12.4/13.3 & 12.8/13.6 & 12.5/13.3 \\ 
    40k & 20.6/22.4 & 12.4/13.3 & 12.8/13.6 & 12.4/13.3 \\ 
    \bottomrule
    
\end{tabular}
\end{table}

Table \ref{tab:sf_vs_lf_training} shows long-form inference and how training utterance length affects accuracy.
In the ``SF" columns, we see that MHA struggles to generalize
when trained on shorter utterances and decoded with longer chunk sizes, while the RWKV-based model
shows
strong generalization.
This suggests that recurrent attention inherently enhances generalization across sequence lengths. 
When trained on longer utterances (the ``LF" columns), both models generalize to sequences far longer than what was seen during training.
For both attention types, performance improves with larger chunk sizes (before plateauing), bolstering the hypothesis that additional context benefits long-form ASR inference.

\begin{table*}[t]
    \caption{Throughput and WER performance of LF-trained and LFXL-fine-tuned models, as a function of decoding chunk size. Throughput reported in minutes of audio processed per second (MPS). Throughput numbers are rounded to the closest minute, for readability.}
  \label{tab:throughput_vs_chunk_acc_LA_round}
  \centering
  {\resizebox{\textwidth}{!}{
  \begin{tabular}{c|cc|ccc|ccc|ccc}
    \toprule
    \multirow{2}{*}{\textbf{chunk}} 
    & \multicolumn{2}{c}{\textbf{MHA} }
    & \multicolumn{2}{c}{\textbf{MHA (LCA256+GT)}} & +FT-LFXL
    & \multicolumn{2}{c}{\textbf{uni-RWKV}} & +FT-LFXL
    & \multicolumn{2}{c}{\textbf{bi-RWKV}} & +FT-LFXL \\
    \cmidrule{2-12}
     & \textbf{MPS} & \textbf{WER (\%)}  
     & \textbf{MPS} & \textbf{WER (\%)}  &\textbf{WER (\%)}
     & \textbf{MPS} & \textbf{WER (\%)}  &\textbf{WER (\%)}
     & \textbf{MPS} & \textbf{WER (\%)}  &\textbf{WER (\%)} \\
    \midrule
      2k &  23 & 13.3/14.1     
         &  10 & 13.3/14.3 & 13.2/14.1        
         &  30 & 13.6/14.4 & 13.5/14.3 
         &  25 & 13.2/14.1 & 13.2/13.9 \\ 
      9k &  13 & 12.5/13.4
         &  17 & 12.6/13.4 & 12.5/13.2   
         &  30 & 12.9/13.5 & 12.7/13.4
         &  25 & 12.6/13.4 & 12.4/13.1 \\
     20k &   7 & 12.4/13.3
         &  18 & 12.5/13.3 & 12.3/13.1   
         &  31 & 13.1/13.6 & 12.4/13.2
         &  26 & 12.5/13.3 & 12.2/13.0\\
     40k &   - & - 
         &  18 & 12.4/13.2 & 12.3/13.1  
         &  31 & 13.0/13.6 & 12.4/13.1
         &  \textbf{26} & 12.4/13.3 & \textbf{12.1/12.9}\\
    \bottomrule
    \bottomrule
  \end{tabular}
    }}
\end{table*}
Table \ref{tab:throughput_vs_chunk_acc_LA_round} gives long-form throughput results, as well as a comparison of our two long-form training techniques. 
With long-form training, the MHA, LCA, and bi-RWKV models achieve virtually the same accuracy: 12.4\%/13.2\% for the LCA model and 12.4\%/13.3\% for MHA and bi-RWKV. 
At this accuracy, the MHA model has a throughput of 7 MPS, the LCA model 18 MPS, and the bi-RWKV model 26 MPS. MHA is the only model whose throughput decreases with longer chunk sizes. The uni-RWKV model has the highest throughput at 31 MPS, but with reduced accuracy relative to the other models.


By performing additional fine-tuning with even longer utterances for one epoch, we can get accuracy improvements for all model types. 
The RWKV models improve more than LCA - likely because the limited context and single global token cannot take full advantage of the long-form training.
After this fine-tuning, uni-RWKV achieves comparable accuracy to LCA+GT, at 72\% higher throughput, while bi-RWKV is more accurate than LCA, at 44\% higher throughput.

We also measured model throughput 
on an AMD EPYC 7R32 CPU. For MHA with LCA+GT it ranges from 0.11 to 0.08 MPS as the chunk size rises from 2k to 20k; while it varies from 0.14 MPS to 0.11 for bi-RWKV.
Despite the much reduced throughput on CPU compared to GPU, bi-RWKV is still more efficient than LCA.

\subsection{Direction Dropout}
\begin{table}[th]
  \caption{Accuracy impact of DirDrop-R2L and DirDrop-Both with various decoding strategies. Decoding strategies: left-to-right (L2R), right-to-left (R2L), alternating direction (Alt) and bidirectional (Bi). WER (\%) on short-form segments.}
  \label{tab:direction_dropout}
  \centering
  \resizebox{\columnwidth}{!}{%
  \begin{tabular}{p{2.3cm}|cccc}
  \toprule
    \multirow{1}{*}{\textbf{Training}} & \multicolumn{4}{c}{\textbf{Decoding}} \\
    &
    \multirow{1}{*}{\textbf{L2R}} & \multirow{1}{*}{\textbf{R2L}}  & \multirow{1}{*}{\textbf{Alt}} & \multirow{1}{*}{\textbf{Bi}} \\
    \midrule
    \raggedright\arraybackslash Uni &  11.2/11.3 & - &-& -  \\ 
    \raggedright \hspace{0.5mm}+ Bi & 14.9/15.0 & 18.4/18.2 & 12.1/12.2 & 11.0/11.0  \\ 
   \raggedright \hspace{1.5mm}+ DirDrop-R2L & 11.7/11.7 & 12.4/12.6 & 11.2/11.3 & 10.9/11.0  \\ 
    \raggedright \hspace{1.5mm}+ DirDrop-Both & 11.8/11.8 & 11.8/11.7 & 11.0/11.2 & 10.9/10.9  \\ 
    \bottomrule
  \end{tabular}
  }
\end{table}

Table \ref{tab:direction_dropout} compares four different model training schemes: Unidirectional left-to-right (Uni), bidirectional (Bi) and bidirectional with two flavors of dropout: DirDrop-R2L and DirDrop-Both.
We then compare these training schemes across different decoding settings (left-to-right, right-to-left, alternating direction and bidirectional). 

Standard bidirectional training produces models that perform catastrophically when used for unidirectional inference, in either direction. With the addition of DirDrop-R2L, the accuracy of the bidirectional model used in L2R mode approaches that of a model trained unidirectionally. 
Training with DirDrop-Both results in equal accuracy for L2R or R2L decoding. 
Further, we observe that alternating direction decoding improves with each added dropout direction, and even outperforms the unidirectionally trained model when trained with DirDrop-Both. 
We also see a small regularizing effect of DirDrop-Both on bidirectional decoding, closing the remaining gap between bidirectional RA and MHA. 

 \begin{table}[th]
   \caption{Decoding strategies for accuracy versus speed trade-off for bi-RWKV model with DirDrop-Both from Table~\ref{tab:direction_dropout}. WER (\%) on short-form segments. MPS @ 40K chunk}
   \label{tab:direction_dropout_mixed}
   \centering
   \begin{tabular}{cccc}
   \toprule
    \textbf{Bi} & \textbf{L2R} & \textbf{Alter} & \textbf{MPS} \\
     \midrule
     None & 11.8/11.8 & 11.0/11.2 & 30 \\
     \midrule
     First      & 11.8/11.8 & 11.0/11.2 & 30 \\ 
     Last       & 11.5/11.6 & 11.1/11.2 & 30 \\ 
     Last Three & 11.3/11.3 & 10.9/11.1 & 29 \\ 
     Last Six   & 11.3/11.2 & 10.9/11.0 & 27 \\ 
     \midrule
     All & \multicolumn{2}{c}{10.9/10.9} & 26 \\
     \bottomrule
   \end{tabular}
 \end{table}

Table~\ref{tab:direction_dropout_mixed} shows the accuracy-speed trade-off enabled by DirDrop.
In particular, we build on the results from Table~\ref{tab:direction_dropout} with the model trained using DirDrop-Both.
Using the DirDrop-Both model, we compare L2R and alternating decoding, each with a few bidirectional layers.
In L2R mode, it is less effective to make the last layer bidirectional than the first one.
For both decoding modes, each additional bidirectional layer improves accuracy while reducing throughput.
Alternating direction decoding is consistently more accurate than L2R decoding and less affected by added bidirectional layers, suggesting it already captures most relevant information from both directions. 

Ultimately, with DirDrop and alternating direction decoding, we can achieve almost the same short-form accuracy as the bidirectional model with only three bidirectional layers, increasing throughput from 26 to 29 MPS. These results show the importance of merging information from the past and future, but reveal that it isn’t necessary to process both directions within each layer. 

\section{Conclusions and Future Work}
In this paper, we showed that RA layers can effectively replace MHA in long-form ASR models, achieving competitive accuracy with significant throughput advantages. Our bi-RWKV-Conformer matches or exceeds MHA and limited-context MHA accuracy, while processing more audio per second. RWKV attention also exhibits strong generalization, maintaining good long-form inference performance even when trained on short utterances. Our novel DirDrop method enables efficient uni- or bi-directional use of the same model. Furthermore, alternating direction decoding is more accurate than L2R decoding at the same throughput and almost as accurate as bidirectional decoding.

Future work includes training long-form models with DirDrop and directly in alternating direction mode. We will also explore extending RWKV to the transducer decoder, enabling cross-layer state passing (as in recent RWKV LLM architectures), and applying RA layers to streaming ASR, where speed and long-range context are crucial.

\pagebreak


\bibliographystyle{IEEEtran}
\bibliography{mybib}

\end{document}